# HiMAL: A Multimodal Hierarchical Multi-task Auxiliary Learning framework for predicting and explaining Alzheimer's disease progression

[1,2]Sayantan Kumar[§], [2]Sean C. Yu, [2,3]Andrew Michelson, [1,2,4]Thomas Kannampallil[†], [1,2]Philip Payne[†], for the Alzheimer's Disease Neuroimaging Initiative

[1] Department of Computer Science and Engineering, McKelvey School of Engineering
[2] Institute for Informatics, Data Science and Biostatistics (I2DB), School of Medicine
[3] Division of Pulmonary and Critical Care, Department of Medicine, School of Medicine
[4] Department of Anesthesiology, School of Medicine

† Contributed equally; co-senior authors

§Corresponding author:

Sayantan Kumar, MS
660 South Euclid Ave, Campus Box 8132
St Louis, Missouri 63110
Sayantan.kumar@wustl.edu
(+1)314.326.9095

**Keywords**: Alzheimer's disease progression; multimodal; multi-task auxiliary learning; hierarchical; model explainability

*Word count*: 3960 words

# ABSTRACT

**Objective**: We aimed to develop and validate a novel multimodal framework HiMAL (**Hi**erarchical, **M**ulti-task **A**uxiliary **L**earning) framework, for predicting cognitive composite functions as auxiliary tasks that estimate the longitudinal risk of transition from Mild Cognitive Impairment (MCI) to Alzheimer's Disease (AD).

**Methods**: HiMAL utilized multimodal longitudinal visit data including imaging features, cognitive assessment scores, and clinical variables from MCI patients in the Alzheimer's Disease Neuroimaging Initiative (ADNI) dataset, to predict at each visit if an MCI patient will progress to AD within the next 6 months. Performance of HiMAL was compared with state-of-the-art single-task and multi-task baselines using area under the receiver operator curve (AUROC) and precision recall curve (AUPRC) metrics. An ablation study was performed to assess the impact of each input modality on model performance. Additionally, longitudinal explanations regarding risk of disease progression were provided to interpret the predicted cognitive decline.

**Results**: Out of 634 MCI patients (mean [IQR] age : 72.8 [67-78], 60% men), 209 (32%) progressed to AD. HiMAL showed better prediction performance compared to all single-modality singe-task baselines (AUROC = 0.923 [0.915-0.937]; AUPRC= 0.623 [0.605-0.644]; all $p<0.05$). Ablation analysis highlighted that imaging and cognition scores with maximum contribution towards prediction of disease progression.

**Discussion**: Clinically informative model explanations anticipate cognitive decline 6 months in advance, aiding clinicians in future disease progression assessment. HiMAL relies on routinely collected EHR variables for proximal (6 months) prediction of AD onset, indicating its translational potential for point-of-care monitoring and managing of high-risk patients.

# BACKGROUND AND SIGNIFICANCE

Alzheimer Disease (AD) is a neurodegenerative disorder characterized by loss of memory and impaired cognition which affects millions of people worldwide.[1–3] AD-related brain pathology, which includes the accumulation and deposition of amyloid-β peptide and tau protein often occurs several years prior to the onset of clinical symptoms, making it challenging to detect AD early.[4,5] Mild cognitive impairment (MCI) is considered a prodromal phase of AD that gradually progresses to AD over several years.[6,7] As only 10% to 20% of MCI patients progress to an AD diagnosis, it is challenging to differentiate between stable MCI (sMCI) patients who do not progress to AD and progressive MCI (pMCI) patients who are develop AD over time.[8–10]

With the widespread applications of machine learning (ML) models in clinical decision support systems, several studies have focused on developing ML-based models for predicting the transition of patients from MCI to AD.[11–15] However, the existing research has three major limitations. First, much of the research has relied on single modality data such as Magnetic Resonance Imaging (MRI), Positron Emission Tomography (PET), clinical data or genetic information.[16–18] However, the causal underpinnings of AD is multifactorial disease requiring multimodal data, with each data modality provides unique and complementary information, cumulatively enabling a more comprehensive understanding of disease pathophysiology and progression.[9,15,19,20] Second, most of the prior models were single-task models, predicting only a specific event such as sMCI vs pMCI or cognitive scores at a future timepoint.[9,21,22] However, at the point-of-care, clinicians have to determine the progression of the disease with concurrent prediction of multiple adverse events such as a future clinical diagnosis and forecasting the progression of cognitive performance over the natural course of the disease.[16,23,24] Finally, most existing research has adopted a cross-sectional prediction approach, relying on baseline data for predicting risk of progression.[9,25]

However, the probability of progression to AD is dynamic, [26] and as such, models need to be flexible to make reliable, longitudinal predictions.

Our objective in this research is to utilize longitudinal data from multimodal electronic health records (EHR) data sources such as imaging, cognition measures, and clinical variables for predicting the longitudinal risk of MCI to AD progression. Towards this end, we had two research objectives: (i) develop a novel hierarchical multi-task deep learning model HiMAL (**Hi**erarchical, **M**ulti-task **A**uxiliary **L**earning), that can predict neuropsychological composite functions as auxiliary tasks to estimate the main task of predicting progression, leading to more informative feature representations and better prediction performance for all tasks, and (ii) provide longitudinal explanations about the potential factors behind disease progression.

# MATERIALS AND METHODS

## Data sources

Data were sourced from the Alzheimer's Disease Neuroimaging Initiative (ADNI) database, established in 2003 through a public-private partnership.[27] ADNI aims to investigate the utility of combining various neuroimaging biomarkers and neuropsychological assessments for tracking the progression of mild cognitive impairment (MCI) and early Alzheimer's disease (AD). All participants in ADNI provided written informed consent, and study protocols were approved by respective institutional review boards at each site. Additional details about ADNI, including study protocols, eligibility criteria, and data access, can be found at http://www.adni-info.org/.

## Cohort selection

We analyzed longitudinal data from ADNI1, ADNI2, ADNI GO, and ADNI3. Participants were selected if diagnosed with 'MCI', 'EMCI' (Early MCI), or 'LMCI' (Late MCI) at baseline, with available T1-weighted MRI scans, cognitive test scores, and clinical data. MCI subjects reverting to Cognitively Normal (CN) during follow-up were excluded, due to the potential uncertainty of the clinical diagnosis, considering that AD is an irreversible form of dementia. MCI subjects progressing to AD within 5 years from baseline visit were labeled as pMCI; others as sMCI.[28] For sMCI, all available imaging and cognition visits within 5 years from baseline were included; for pMCI, visits until progression were selected. All cognition visits were aligned with imaging visits. Note that clinical variables were extracted only from baseline visits for both sMCI and pMCI samples. The cohort was split into training, validation, and test sets in a ratio of 75:15:10 for model training, hyperparameter tuning, and evaluation, respectively.

## Variables and data processing

### Multimodal input feature selection

Imaging data used in our analysis included preprocessed MRI brain region volumes (gray matter volumes) obtained from T1-weighted MRI scans. We included 90 imaging features including volumes of 66 cortical regions (33 per hemisphere) and 24 subcortical regions (12 per hemisphere). Further details about how the regional brain volumes are extracted from MRI scans be found both online (https://adni.loni.usc.edu/methods/mri-tool/mri-analysis/) [29] and in the Supplementary Material S1.1. Following existing studies on AD progression, we included 13 neuropsychological cognitive assessment scores and 18 clinical features including demographics of participants (age, sex), vital signs and medical history (comorbidities). More information about cognitive and clinical features, along with the full list of features can be found in the Supplementary Material (S.1.2 and S.1.3).

### Composite cognitive functions as auxiliary tasks

We used four neuropsychological composite cognitive functions as our auxiliary tasks which have been validated to have associated with the risk of progressing from MCI to AD. These included composite scores for different domains of cognitive performance, namely memory (ADNI-MEM),[30] composite score for executive functioning (ADNI-EF),[31] composite score for language cognitive terms (ADNI-LAN) and composite score for visual-spatial (ADNI-VS).[32] Further, the use of a composite score increased measurement precision, helps avoid idiosyncratic features of a particular test that may capitalize on chance, and limits the number of statistical tests needed compared to analyzing each of the constituent parts separately. Details of how each of the composites were calculated from individual neuropsychological test batteries can be found in the Supplementary Material S2.

### Data preprocessing

All imaging features of each patient were normalized by the intracranial volume (ICV) of that patient. Feature pre-processing of continuous time-series variables (imaging, cognition measures and composite functions) included clipping the outlier values to the 5th and 95th percentile values and scaling between [0,1] using the MinMaxScalar package from sklearn. The training set was first scaled and the parameters from the scaled training set were used to standardize the validation and test set respectively. All categorical features were one-hot encoded.

### Missing value imputation

Longitudinal data for each participant were collected at an interval of 6 months to 5 years from the baseline visit. However, not all participants had all modalities recorded at every visit.

Missing values for all continuous time-series variables were imputed using the BRITS algorithm, a state-of-the-art imputation based on recurrent dynamics.[33] Missing values for categorical variables were filled using the mode value of each variable. Following prevalent techniques in AD literature,[34] diagnosis information (MCI or AD) for each participant was forward filled from their previous recorded visit, if missing.

## Ground truth labels for clinical outcome and composite cognitive scores

For the primary task, we aimed to predict the risk of AD progression at each timepoint within the visit trajectory of an individual. Specifically, HiMAL predicted at each timepoint(visit) the probability (risk) of a subject progressing to AD by their next visit (within 6 months from current visit). The ground truth labels for the primary task were calculated as follows (Figure 1). At each timepoint, if the diagnosis label is AD for the next visit, we labelled that visit with outcome 1 or else 0. For example, if a pMCI subject progresses after 24 months from the baseline visit, the labels for bl (baseline), m06 (month 6), m12 (month 12) will be 0; while the labels for m18 (month 18) and m24 (month 24) will be 1 (Figure 1). Similar to the primary task, the predicted auxiliary tasks were also anticipated longitudinal predictions. Specifically, HiMAL predicted at each timepoint (visit) the forecasted composite scores at the next visit or after 6 months (Figure 1).

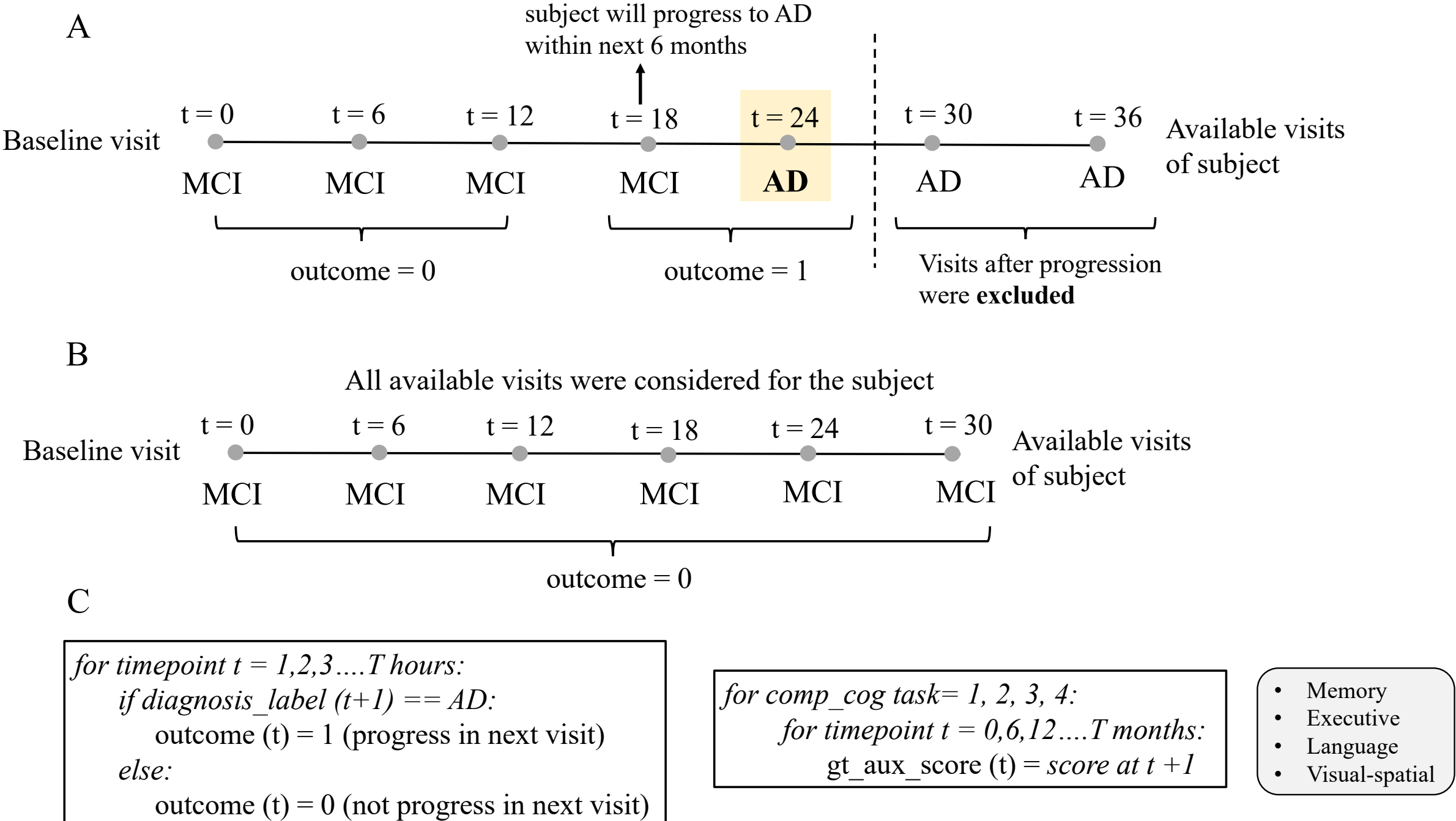

**Figure 1**: Ground-truth label for clinical outcome (main task) and composite cognitive scores (auxiliary tasks) mortality. **(A)** Example pMCI subject who had available visits till 3 years from baseline (t = 36) and progressed at t = 24. Outcome labels were 0 for t = 0 until t = 12 and 1 for t = 18 (progress at next visit) and t = 24. Visits after progression (t = 24 onwards) were excluded. (B) Example sMCI who never progressed. All available visits till t = 30 were considered with all visits marked outcome label 0. (C) Pseudo-code showing how ground truth labels are calculated for prediction of progression (left) and the 4 composite cognitive scores (right).

## Hierarchical multi-task auxiliary learning framework

Our proposed hierarchical multi-task auxiliary learning framework (HiMAL) can be divided into three main components (Figure 3): (i) multimodal feature embedding with time-series embedding, (ii) hierarchical task predictions and (iii) model explanations (Figure 2).

### Multimodal feature embedding with time-series imputation

HiMAL takes baseline clinical data, longitudinal MRI gray matter volumes (imaging data) and longitudinal cognitive assessment scores as the multimodal input. The time-invariant clinical features are passed through a set of fully connected layers to obtain the clinical embedding. Similarly, the multivariate time-series features (imaging and cognition data) are passed through a recurrent module to estimate the corresponding imaging and cognition embeddings respectively. Note that the recurrent module consists of a series of Long Short-Term Memory (LSTM)[35] layers along with provisions for missing value imputation using the BRITS algorithm. BRITS is a data-driven imputation procedure where the missing values were imputed based on recurrent dynamics.[33] More details about missing value imputation can be found in Supplementary Material S3. Next, the embeddings from the individual modalities were concatenated to obtain the joint embedding space (Figure 2).

**Hierarchical task predictions**

The concatenated multimodal feature embedding was subsequently passed through an auxiliary network of fully connected layers followed by a sigmoid activation function to predict the memory, executive functioning, language, and visual-spatial composite scores as the four auxiliary tasks (Figure 2). The main task of predicting MCI to AD progression was estimated by a weighted combination of the auxiliary tasks. Specifically, the predicted composite scores were multiplied with the relevance weights and passed through a fully connected layer, followed by a sigmoid activation function to predict the risk of progression. The relevance weights were learnt by passing the concatenated embedding through relevance network of fully connected layers followed by a sigmoid activation function. Similar to the main and auxiliary tasks, the relevance weights were also predicted for every visit timepoint of each participant in our cohort (Figure 2).

HiMAL differs from existing multi-task learning methods where the primary and auxiliary tasks were predicted concurrently without any hierarchical dependency between them. The hierarchical nature of HiMAL allowed the model to exploit task dependencies more effectively, leading to more informative feature representations and effective knowledge transfer between the tasks.

## Model explanations

The relevance weights represented the contribution (importance) of each forecasted composite score towards the main task of predicting progression. Since progression prediction was calculated as a function of the forecasted composite scores at each timepoint, we aimed to analyze which of the composite scores are important predictors of progression from MCI to AD. The relevance weights and the forecasted composite scores at each timepoint provided insights about how the contributions of different composites change over time from baseline visit to the onset of progression. For example, HiMAL can provide an explanation (at a particular timepoint) that the participant will probably progress to AD in the next 6 months due to possible decline in memory and executive functioning. Since the weights were jointly optimized along with the main and auxiliary tasks, the explanations were provided along with prediction within the same end-to-end framework, without relying on post-hoc explanation techniques (Figure 2).

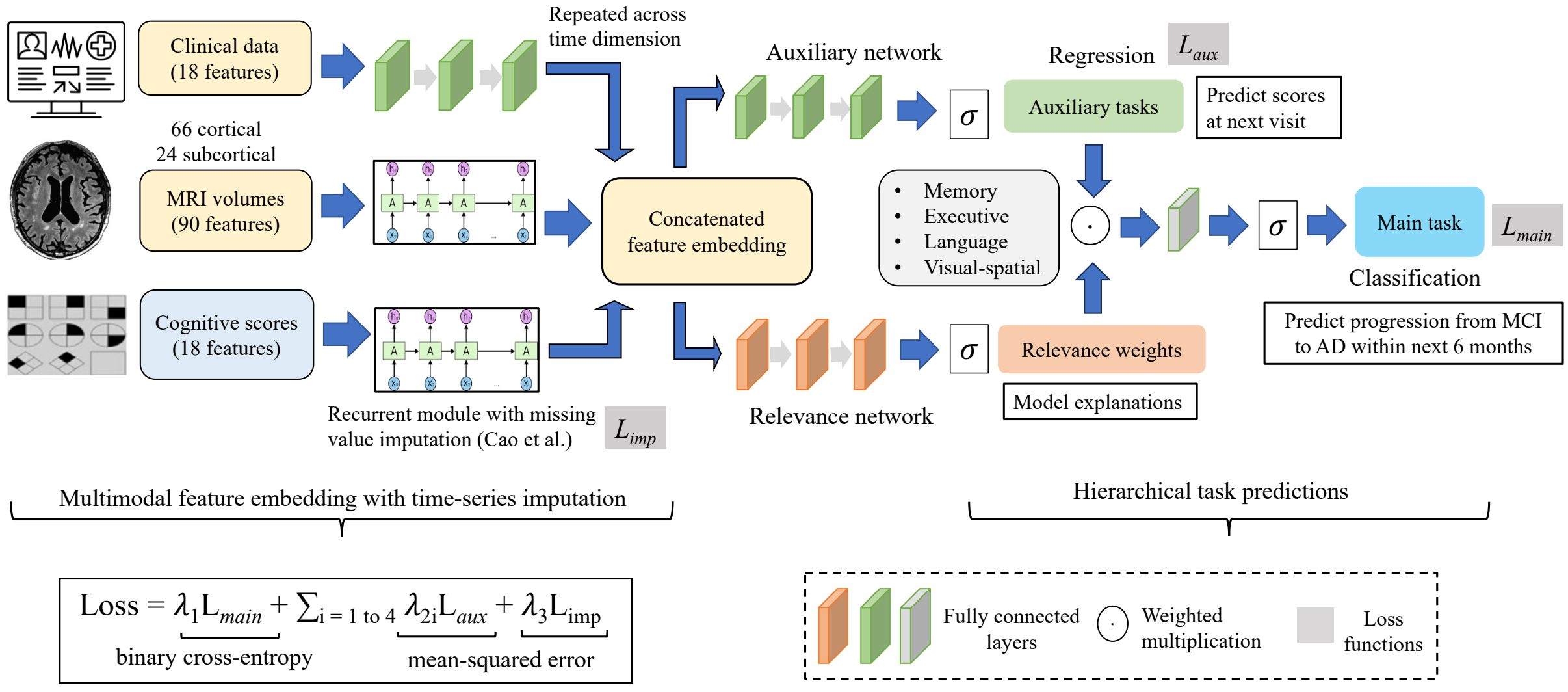

**Figure 2**: HiMAL: multimodal **Hi**erarchical **M**ulti-task **A**uxiliary **L**earning framework for predicting risk of MCI to AD progression using forecasted cognitive composite scores. Time-series input features (imaging and cognition data) passed through recurrent modules with missing value imputation and time-invariant clinical data passed through a series of fully connected layers were concatenated into a multimodal feature embedding. The concatenated feature embedding was passed through two separate networks to predict the auxiliary tasks and the corresponding relevance weights at each timepoint. The main task was estimated by the weighted combination of the auxiliary scores and the relevance weights followed by a fully connected layer and sigmoid activation function.

## Model training and evaluation

### Loss function

The loss function of HiMAL can be written as: $L = \lambda_1 L_{main} + \lambda_2 \sum_{c=1}^{4} L_{aux} + \lambda_3 L_{imp}$ where $L_{main}$ and $L_{aux}$ are the binary cross-entropy loss for the main task classification and the mean-squared error for the auxiliary task regression respectively. The imputation loss $L_{imp}$ was also calculated using the mean-squared difference between the original and imputed values (for details, see [33] and Supplementary Material S3). $\lambda_1$, $\lambda_2$ and $\lambda_3$ represent the coefficients of the different losses respectively and were set to $\lambda_1$= 1, $\lambda_2$ = 0.8 and $\lambda_3$ = 0.05 respectively after hyperparameter tuning using grid search on the validation set.

### Comparison with baselines

We compared the performance of HiMAL with the following baseline models (Figure 3): (i) single-modality single task models (SMST)[9], (ii) single-modality multi-task models (SMMT)[36], (iii) multimodal single task models (MMST)[37] and (iv) multimodal multitask models (MMMT).[16] Note that all these baseline models are well-validated benchmarks for AD progression prediction. For models with a single modality (SMMT, SMST), the input can be either of the 3 modalities (imaging only, cognition only and clinical only). The models having multiple tasks (SMMT, MMMT), predicted both the main and auxiliary tasks in a traditional multi-task auxiliary learning approach where the main and auxiliary tasks are predicted independently in parallel (Figure 3). We also included traditional machine learning techniques like Support Vector Machines (SVM)[38], Random Forests (RF)[39], and XGBoost[40] as baselines for comparison. These models were inherently cross-sectional and treated the time-series data as independent data points. These baselines received multiple modalities concatenated as a single input to predict progression from MCI to AD.

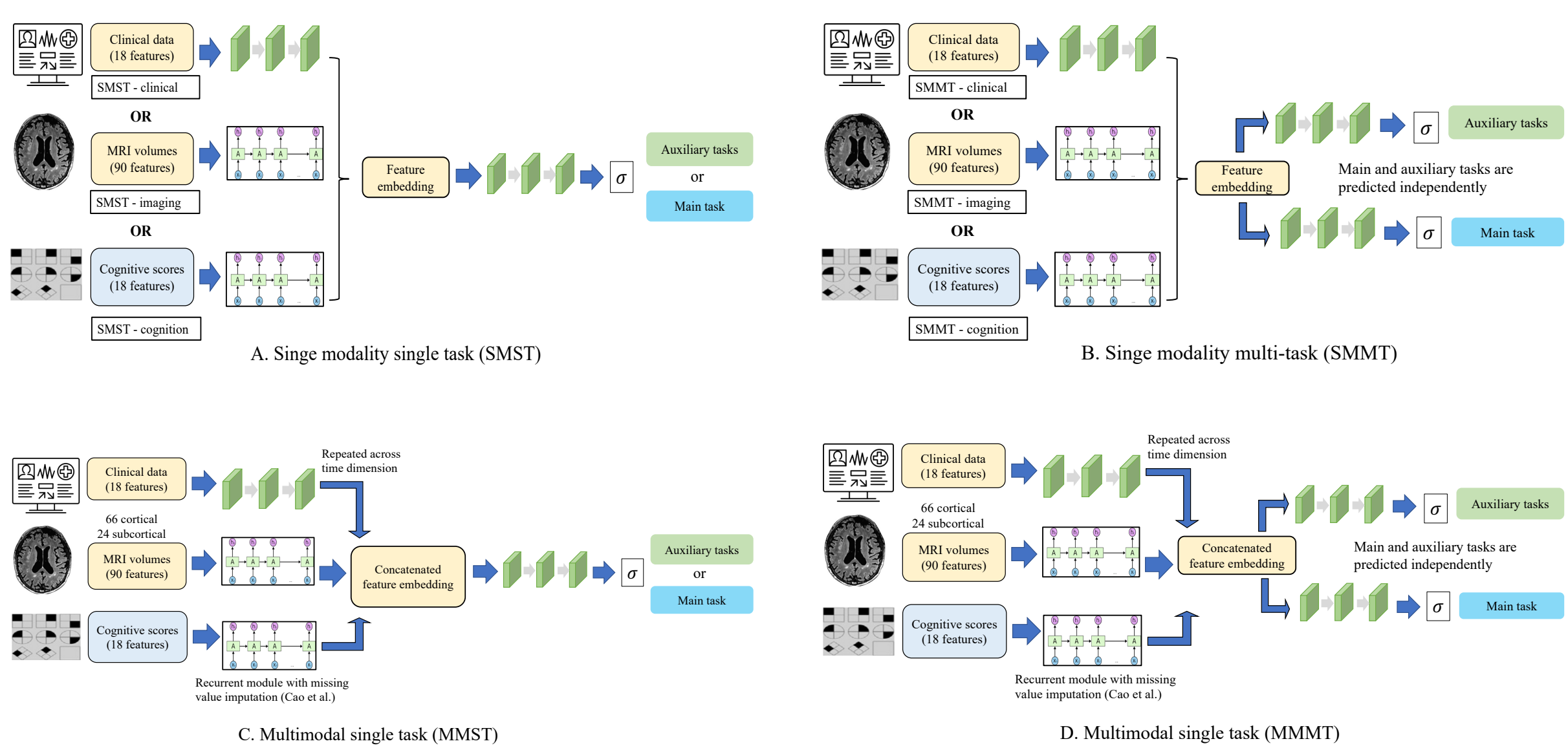

**Figure 3**: Baseline models used as comparison in our study. (A) Single modality single task model (SMST), (B) single modality multi-task (SMMT) model, (C) multimodal single-task model (MMST) and multimodal multi-task model (MMMT). For SMMT and MMMT, the main and auxiliary tasks were predicted independently.

## Statistical analysis

All models (HiMAL and baselines) were trained with the same set of multimodal features and network parameters. Details of the parameter configurations are available in Supplementary Material S4. We reported both the Area Under the Receiver Operating Characteristics (AUROC)[41] and the Area Under the Precision Recall Curve (AUPRC)[42] for the main task of progression prediction and Mean Squared Error (MSE) for the composite score regression. For all reported results, we computed the 95% confidence intervals (CIs) using the pivot bootstrap estimator[43] by sampling participants from the test dataset with replacement 200 times. We also performed the 2-sided Wilcoxon signed rank tests[44] to pairwise compare HiMAL and each of the baseline methods for the main task. The critical value $p = 0.05$ was selected and the final p values were reported after false discovery rate (FDR) correction to adjust for multiple hypotheses.[45]. We also performed an ablation study to understand the impact of each input modality on the performance of HiMAL. Specifically, the performance metrics on the main task (AUROC/AUPRC) and the auxiliary tasks (MSE) were compared for different combination of the 3 modalities.

## RESULTS

Figure 4 shows a flowchart of cohort selection including the inclusion and exclusion criteria. We first identified 1091 MCI subjects out of which 725 subjects were selected with simultaneous availability of T1-weighted MRI scan, neuropsychological cognitive test scores, and clinical data at the baseline visit. 91 subjects, who were clinically diagnosed as MCI at baseline visit but reverted to CN during follow-up were excluded from the study. Our final cohort consisted of 425 sMCI samples and 209 pMCI samples. The training, validation and test set included 475, 95 and 64 subjects respectively.

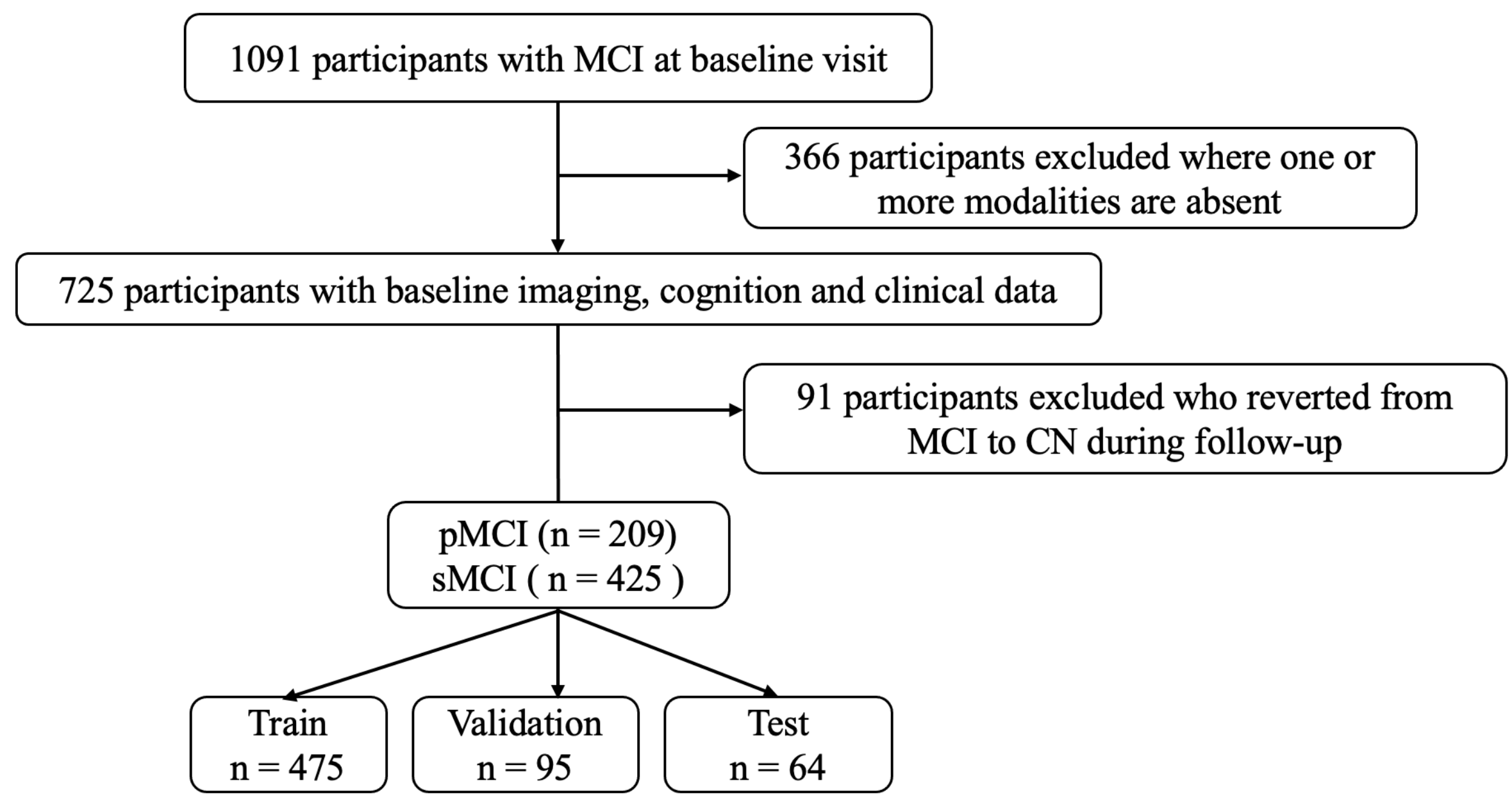

**Figure 4**: Flowchart of study participants. Abbreviations: CN = cognitively normal, MCI = mild cognitive impairment, pMCI = progressive MCI, sMCI = stable MCI.

## Sample characteristics

Descriptive statistics corresponding to the baseline visits for pMCI and sMCI cohorts respectively are shown in Table 1. Our cohort consisted of more MCI individuals who remained stable in their trajectory (sMCI; n = 425) compared to those who progressed (pMCI; n = 209). The sMCI cohort had relatively younger individuals (mean +/- STD: [72.5 +/- 5.4] vs [73.7 +/- 5.26]; p = 0.037) and higher proportion of males (252/425 = 59% vs 115/209 = 55%; p = 0.007) compared to the pMCI cohort. The level of education (in years) was slightly higher for the pMCI cohort (mean +/- STD:  [16.2 +/- 2.55] vs [15.4 +/- 2.27]), though the difference was not statistically significant. The pMCI cohort had less AD pathology than sMCI indicated by the generally high MMSE scores (mean +/- STD : [28.7 +/- 1.7] vs [24.8 +/- 2.1]; p = 0.0035). In terms of the composite cognitive scores, pMCI individuals exhibited more impairment (lower values) in memory ((mean +/- STD : [-0.98 +/- 0.68] vs [0.52 +/- 0.81]; p < 0.001), executive functioning ((mean +/- STD : [1.1 +/- 0.85] vs [0.53 +/- 0.76]; p < 0.001), language ((mean +/- STD : [-1.01 +/- 1.1] vs [0.48 +/- 0.82]; p = 0.008) and visual-spatial skills ((mean +/- STD : [-0.76 +/- 1.07] vs [0.07 +/- 0.74]; p < 0.001) compared to the sMCI cohort.

**Table 1**: Descriptive statistics of the pMCI and sMCI cohorts respectively. All variables corresponded to the baseline visit. [a]: Range from [0,30] higher scores indicating more impairment.[b]: Range from [–3,3] with lower values indicating more impairment. [c]: Range from [–3,1] with lower values indicating more impairment. **Abbreviations**: STD = Standard deviation, MMSE = Mini-Mental State Examination.

| Variables | Total cohort | pMCI | sMCI |
|---|---|---|---|
| Number of samples, n (%) | 634 (100%) | 209 (33%) | 425 (67%) |
| Age (years), mean +/- STD | 72.8 +/- 4.75 | 73.7 +/- 5.26 | 72.5 +/- 5.4 |
| Male: Female | 375:259 | 115:94 | 252:173 |
| Education (years), mean +/- STD | 15.9 +/- 2.76 | 15.4 +/- 2.27 | 16.2 +/- 2.55 |

| [a] MMSE, mean +/- STD | 27.5 +/- 1.9 | 24.8 +/- 2.1 | 28.7 +/- 1.7 |
|---|---|---|---|
| [b] Memory, mean +/- STD | 0.13 +/- 0.93 | -0.98 +/- 0.68 | 0.52 +/- 0.81 |
| [b] Executive, mean +/- STD | 0.125 +/- 1.07 | -1.1 +/- 0.85 | 0.53 +/- 0.76 |
| [b] Language, mean +/- STD | 0.07 +/- 0.91 | -1.01 +/- 1.1 | 0.48 +/- 0.82 |
| [c] Visual-spatial, mean +/- STD | -0.12 +/- 0.85 | -0.76 +/- 1.07 | 0.07 +/- 0.74 |

## Model performance

HiMAL outperformed all baseline methods including traditional machine learning classifiers, single task and unimodal baselines in terms of both AUROC and AUPRC (Table 2). The non-deep learning models SVM, RF and XGBoost were cross-sectional ignoring the temporal aspect of the data, and hence exhibited a general trend of lower prediction performance compared to their deep learning counterparts. Overall, we observed that using multiple modalities is beneficial compared to using only a single modality as demonstrated by the superior performance of MMST and MMMT over SMST and SMMT. For a particular modality configuration (e.g. SMST vs SMMT or MMST vs MMMT), multi-task models performed better than their single-task counterparts.

We observed similar trends of HiMAL performing better (lower mean squared error) than the unimodal and single-task baselines for the auxiliary task regression (Table 3). HiMAL had the lowest MSE for memory and language and were second best for executive functioning and visual-spatial. Overall, multimodal models exhibited better performance for all the composite cognitive scores compared to their unimodal counterparts as demonstrated by the superior performance of MMST and MMMT over SMST and SMMT respectively (Table 3). We observed a general trend of multi-task models performing better than single-task models, which was consistent across all the four composite scores.

**Table 2**: Performance on the main task (AD progression prediction) in terms of AUROC (higher = better) and AUPRC (higher = better) for HiMAL and the baselines. The numbers in [ ] indicate 95% confidence intervals using the pivot bootstrap estimator by sampling participants from the test dataset with replacement 200 times. The p-values for AUROC and AUPRC were calculated by pairwise comparison between the proposed (last row) and each of the baselines using the Wilcoxon signed rank tests adjusted for False Discovery Rate (FDR) correction. Significant p-values are highlighted in bold with *: 0.01 < p < 0.05, **: 0.005 < p < 0.01, ***: p < 0.001. **Abbreviations**: SVM = Support Vector Machines, RF = Random Forest, SMST = Single-modality single task, SMMT = single modality multi-task, MMST = multimodal single task, MMMT = multimodal multi-task

| Model | AUROC | *p-value (AUROC) | AUPRC | *p-value (AUPRC) |
|---|---|---|---|---|
| SVM | 0.714 [0.705-0.728] | **< 0.001*****| 0.416 [0.405-0.431] | **< 0.001***** |
| RF | 0.688 [0.67-0.695] | **< 0.001***** | 0.392 [0.377-0.408] | **< 0.001***** |
| XGBoost | 0.725 [0.708-0.738] | **< 0.001***** | 0.436 [0.422-0.451] | **< 0.001***** |
| SMST | | | | |
| - Imaging only | 0.747 [0.725-0.758] | **< 0.001***** | 0.432 [0.415-0.452] | **< 0.001***** |
| - Cognition only | 0.715 [0.726-0.735] | **< 0.001***** | 0.429 [0.421-0.437] | **< 0.001***** |
| - Clinical only | 0.637 [0.615-0.656] | **< 0.001***** | 0.365 [0.296-0.421] | **< 0.001***** |
| SMMT | | | | |
| - Imaging only | 0.894 [0.865-0.918] | **0.018*** | 0.542 [0.517-0.568] | **< 0.001***** |
| - Cognition only | 0.825 [0.805-0.847] | **< 0.001***** | 0.472 [0.452-0.481] | **< 0.001***** |
| - Clinical only | 0.723 [0.708-0.734] | **< 0.001***** | 0.443 [0.422-0.461] | **< 0.001***** |
| MMST | 0.875 [0.862-0.898] | **< 0.001***** | 0.574 [0.507-0.595] | **0.0027**** |
| MMMT | 0.902 [0.875-0.933] | **0.037*** | 0.595 [0.537-0.625] | **0.0085**** |
| **HiMAL** | **0.923 [0.915-0.937]** | - | **0.623 [0.605-0.644]** | - |

**Table 3**: Performance on the auxiliary task regression (memory, executive functioning, language and visual-spatial) in terms of Mean squared error (MSE; lower = better) for HiMAL and the baselines. The numbers in [ ] indicate 95% confidence intervals using the pivot bootstrap estimator by sampling participants from the test dataset with replacement 200 times. **Abbreviations**: SMST = Single-modality single task, SMMT = single modality multi-task, MMST = multimodal single task, MMMT = multimodal multi-task

| | **Regression (Mean squared error)** | | | |
|---|---|---|---|---|
| **Model** | **Memory** | **Executive** | **Language** | **Visual-spatial** |
| SMST | | | | |
| - Imaging only | 0.352 [0.332-0.367] | 0.376 [0.362-0.385] | 0.406 [0.376-0.428] | 0.416[0.435-0.468] |
| - Cognition only | 0.395 [0.365-0.412] | 0.358 [0.334-0.375] | 0.352 [0.341-0.368] | 0.416 [0.405-0.44] |
| - Clinical only | 0.526 [0.508-0.551] | 0.516 [0.507-0.528] | 0.62 [0.606-0.651] | 0.575 [0.542-0.594] |
| SMMT | | | | |
| - Imaging only | 0.234 [0.195-0.252] | 0.224 [0.212-0.245] | 0.27 [0.256-0.296] | 0.295 [0.276-0.307] |
| - Cognition only | 0.345 [0.292-0.363] | 0.326 [0.312-0.34] | 0.334 [0.315-0.347] | 0.305 [0.286-0.337] |
| - Clinical only | 0.422 [0.391-0.456] | 0.435 [0.423-0.45] | 0.387 [0.362-0.403] | 0.406 [0.392-0.418] |
| MMST | 0.185 [0.158-0.207] | 0.206 [0.188-0.221] | 0.24 [0.23-0.256] | 0.245 [0.228-0.267] |
| MMMT | 0.21 [0.19-0.232] | **0.175 [0.152-0.188]** | 0.252 [0.232-0.265] | **0.216 [0.195 – 0.238]** |
| **HiMAL** | **0.182 [0.162-0.195]** | 0.195 [0.17-0.225] | **0.232 [0.213-0.256]** | 0.225 [0.218 – 0.247] |

## Model explanations

The forecasted composite scores and the corresponding relevance scores at each timepoint provided insights (explanations) about how the contributions of different composites change over time from baseline visit to the onset of progression. Here we visualized the longitudinal explanations (every 6 months) for a particular pMCI individual who progressed 48 hours after the 1$^{st}$ baseline visit (Figure 5). At each timestep (visit), the outcome (progression) labels from t=0 to t= 36 were 0 (not progress) and 1 (progress) for t=42 and t=48. As the predicted probability of progression rises from t = 36, the model was shown to pay more relevance to memory and language composites (Figure 5). In other words, HiMAL predicted that the MCI patient will progress to AD within the next 6 months because due to forecasted decline cognitive skills related to memory and language (Figure 5). Low relevance scores assigned to executive functioning and visual-spatial can be attributed to the fact that these scores were relatively constant with negligible changes throughout the trajectory (Figure 5).

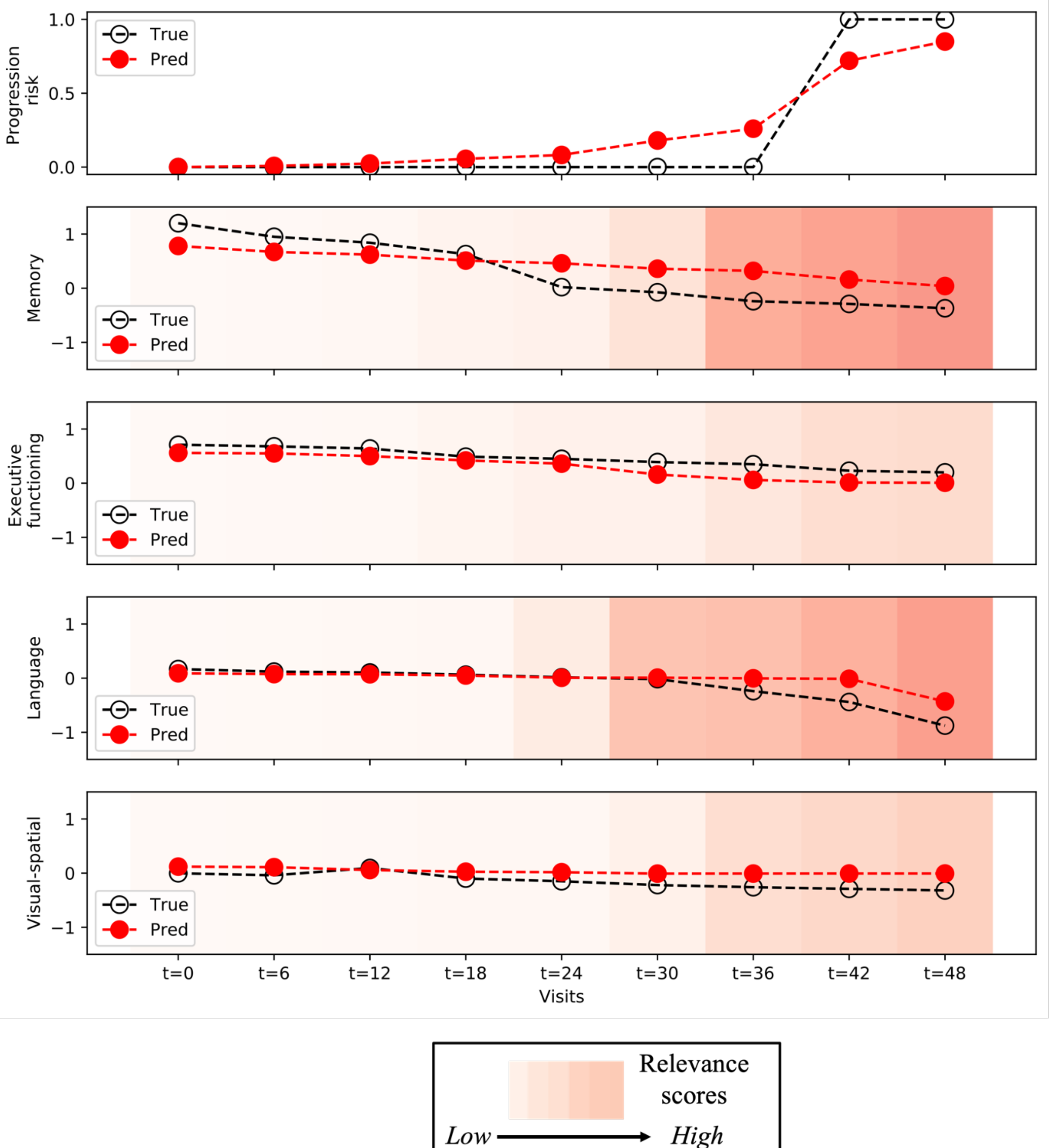

**Figure 5**: Model explanations of an pMCI individual who progressed 48 months (4 years) after the baseline visit. The topmost plot represents progression risk while the following points represent the composite functions. The x-axis represents visits in months and the plotted values represent either the ground truth label (black) or the predicted value (red).The colormap within each row indicates the weight/importance (relevance score) assigned to the forecasted composite scores where dark hues corresponded to higher relevance given to a particular composite score at a specific time point. Relevance scores were scaled between 0 and 1 for easier interpretation. Composite scores, predicted between [0,1] after sigmoid activation function (Figure 3), were rescaled to Z-scores with lower values indicating greater impairment.

## Ablation studies on the number of modalities (HiMAL)

Using all 3 modalities (imaging + cognition + clinical) as input resulted in the best prediction performance of HiMAL for both AD progression prediction (AUROC, AUPRC) and composite cognitive score regression (MSE) (Figure 6). Comparing across modalities, using only imaging data or cognition data as input led to a competitive performance across the different tasks. Combining both imaging and cognition data as bimodal input often resulted in improved performance compared to using either modality as standalone input. This was observed particularly for AUROC (Figure 6A), memory (Figure 6C), and visual-spatial (Figure 6F). However, using only clinical data showed a loss in prediction performance, as demonstrated by the relatively lower AUROC, AUPRC and relatively higher MSE for the composite scores. Combining clinical data as an additional modality to either imaging or cognition features did not result in a significant performance improvement compared to using only imaging or only cognition as unimodal input.

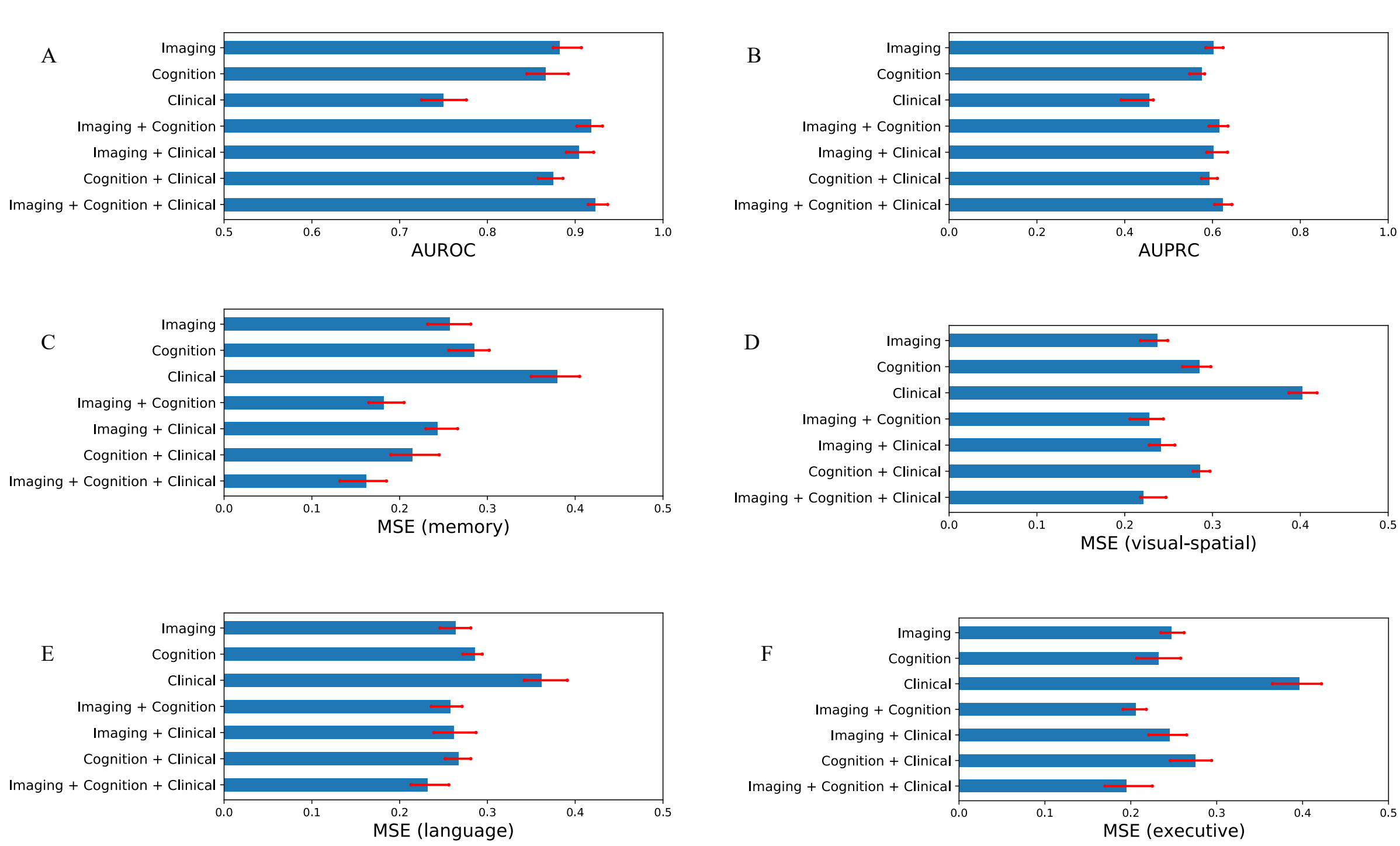

**Figure 6**: Performance of HiMAL for different combinations of input modalities in terms of (A) AUROC, (B) AUPRC, (C) MSE (memory), (D) MSE (executive), (E) MSE (language) and (F) visual-spatial. The red bars represent the 95% confidence interval using the pivot bootstrap estimator by sampling participants from the test dataset with replacement 200 times.

# DISCUSSION

In this study, we used longitudinal multimodal data on imaging, self-reported cognition measures, and clinical data for predicting the longitudinal risk of MCI to AD progression within the next 6 months. We developed and validated HiMAL: a multimodal hierarchical multi-task auxiliary learning framework to predict every 6 months, a set of neuropsychological composite cognitive function scores. Next, we used the forecasted composite function scores at every timepoint to predict the future risk of progression from MCI to AD. Experiments were conducted on the publicly available ADNI dataset, and the findings showed that HiMAL performed better than state-of-the-art unimodal and single-task baselines for predicting both the risk of AD progression (Table 1) and the composite scores (Table 2). Further, HiMAL also provided explanations about the potential factors behind disease progression by analyzing the contributions of each auxiliary task (Figure 5). Ablation analysis revealed imaging and cognition data as the most influential modalities towards the outcome.

For the primary task of predicting risk of MCI to AD progression, multi-task models performed better than their single-task counterparts (e.g. SMST vs SMMT or MMST vs MMMT; Table 2). Additionally, optimizing only a particular composite score did not lead to a drop in performance (MSE) for that score compared to when all the scores are optimized together (MMST vs MMMT vs proposed; Table 3). Both these observations highlighted the benefits of jointly optimizing multiple tasks compared to learning a single task. In terms of architectural design, HiMAL was closest to MMMT where the main and auxiliary task were predicted concurrently without any hierarchical dependance between the tasks. The superior prediction performance of HiMAL on both primary and auxiliary tasks (Table 2 and Table 3) validated our design choice of having a hierarchical relation between the tasks. The superior performance of HiMAL can be attributed to the fact that the hierarchical relationship allowed the model to exploit task dependencies more effectively, leading to more informative feature representations and effective knowledge transfer between the tasks.

There exists a plethora of studies predicting the risk of MCI to AD progression at a single endpoint.[8,9,46] However, the probability of progression from MCI to AD is dynamic and should be assessed using a longitudinal lens. Further, at the point-of-care, clinicians must determine the progression of the disease with concurrent prediction of multiple adverse events such as a future clinical diagnosis and forecasting the progression of cognitive performance over the natural course of the disease. HiMAL provided a more holistic approach for monitoring the risk of progression over a 6-month period, with simultaneous prediction of the functional decline in different domains of human cognition. Further, HiMAL is built on routinely collected EHR variables from a publicly available multimodal AD dataset, which encourages the translational ability of our work on other AD datasets. We believe that HiMAL can be a useful tool for early identification of patients in the prodromal phase of AD, who are at risk of progressing in future. This could potentially lead to therapeutic interventions to delay the disease progression over time and could be helpful for tailoring disease management and planning future care.

In addition to forecasting risk of disease progression, HiMAL also provided a complementary approach of providing explanatory support to the predictions. As black-box deep learning models become more prevalent in healthcare, clinicians seek not only superior prediction performance but also explanations for model decisions to enhance interpretability and trustworthiness.[47–49] Recent research on AD progression only identified the important features for a prediction, providing a snapshot of model explanations at a fixed timepoint.[47,50] On the contrary, HiMAL provided longitudinal explanations about the potential reasons behind high/low risk prediction throughout the visit trajectory. The longitudinal explanations are clinically informative in the sense that they can inform clinicians 6 months in advance the potential cognitive function decline that can lead to progression to AD in future (Figure 5).

HiMAL is also flexible with the selection of modalities and auxiliary, encouraging reproducibility on other clinical problems besides AD. First, our framework can easily accommodate additional modalities with modality-specific modules to generate the feature embedding as shown in Figure 2. Finally, the composite cognitive functions as auxiliary tasks were calculated from individual cognitive assessment scores which are readily available in most AD datasets. The rationale behind the design choice of selecting composite cognitive functions as auxiliary tasks was that the aggregated summary scores represented a particular cognitive domain as a whole and facilitated clinically intuitive units of explanation for analysis of disease progression. However, we would like to emphasize that HiMAL does not depend on the availability of these composite cognitive functions. Other scores like MMSE, ADAS, and CDR-SB can serve as suitable alternatives for auxiliary tasks if necessary.

**Limitations and scope for future work**

Our study has several limitations and opportunities for future research. Firstly, our findings are based on a single AD dataset from North America, limiting generalizability. Future studies should validate our approach on diverse cohorts such as OASIS-3,[51] AIBL[52] and MIRIAD.[53] Secondly, HiMAL only incorporates imaging, cognition, and clinical data, excluding other modalities like PET imaging and genetic data commonly used in AD research.[54] However, HiMAL can accommodate additional modalities, provided there is sufficient longitudinal data and sample size. Thirdly, HiMAL was developed retrospectively, necessitating evaluation in a prospective setting where patient outcomes are unknown. Finally, we emphasize that this are prototype models to demonstrate methods. In order to translate these models into deployment, more rigorous evaluation would be needed including prospective validation and detailed case review.

## CONCLUSIONS

In this study, we utilized longitudinal data from MRI, cognition, and clinical sources to predict the risk of MCI to AD progression. Our hierarchical multi-task deep framework HiMAL predicted neuropsychological composite functions as auxiliary tasks at each timepoint to estimate progression risk. Results on an AD dataset showed superior performance compared to unimodal and single-task approaches. Ablation analysis revealed imaging and cognition data as the most influential modalities towards the outcome. Clinically informative model explanations anticipate cognitive decline 6 months in advance, aiding clinicians in future AD progression assessment. Flexibility in input modalities and auxiliary task selection indicated the translational impact of HiMAL extends to diverse clinical prediction problems.

## FUNDING

The preparation of this report was supported by the Centene Corporation contract (P19-00559) for the Washington University-Centene ARCH Personalized Medicine Initiative.

## AUTHOR CONTRIBUTIONS

All authors contributed to the design of the methodology and the experiments. SK implemented the data preprocessing, modeling, and data analysis and manuscript writing. SY provided guidance on data analysis, model building, and validation. AM, TK and PROP provided conceptualization, clinical interpretation, and manuscript writing. All authors participated in manuscript revision and editing and approved the final version for submission.

## SUPPLEMENTARY MATERIAL

A supplementary file has been submitted along with the main manuscript.

## ACKNOWLEDGEMENTS

Data collection and sharing for this project was funded by the Alzheimer's Disease Neuroimaging Initiative (ADNI) (National Institutes of Health Grant U01 AG024904) and DOD ADNI (Department of Defense award number W81XWH-12-2-0012). ADNI is funded by the National Institute on Aging, the National Institute of Biomedical Imaging and Bioengineering, and through generous contributions from the following: AbbVie, Alzheimer's Association; Alzheimer's Drug Discovery Foundation; Araclon Biotech; BioClinica, Inc.; Biogen; Bristol-Myers Squibb Company; CereSpir, Inc.; Cogstate; Eisai Inc.; Elan Pharmaceuticals, Inc.; Eli Lilly and Company; EuroImmun; F. Hoffmann-La Roche Ltd and its affiliated company Genentech, Inc.; Fujirebio; GE Healthcare; IXICO Ltd.; Janssen Alzheimer Immunotherapy Research & Development, LLC.; Johnson & Johnson Pharmaceutical Research & Development LLC.; Lumosity; Lundbeck; Merck & Co., Inc.; Meso Scale Diagnostics, LLC.; NeuroRx Research; Neurotrack Technologies; Novartis Pharmaceuticals Corporation; Pfizer Inc.; Piramal Imaging; Servier; Takeda Pharmaceutical Company; and Transition Therapeutics. The Canadian Institutes of Health Research is providing funds to support ADNI clinical sites in Canada. Private sector contributions are facilitated by the Foundation for the National Institutes of Health (www.fnih.org). The grantee organization is the Northern California Institute for Research and Education, and the study is coordinated by the Alzheimer's Therapeutic Research Institute at the University of Southern California. ADNI data are disseminated by the Laboratory for Neuro Imaging at the University of Southern California.

## CONFLICTS OF INTEREST

The authors do not have any competing interests to disclose.

## DATA AVAILABILITY

ADNI data used in this study are publicly available and can be requested following ADNI Data Sharing and Publications Committee guidelines: https://adni.loni.usc.edu/data-samples/access-data/.